\documentclass[letterpaper]{article} %
\usepackage[preprint]{aaai2027}  %
\usepackage[hyphens]{url}  %
\usepackage{graphicx} %
\usepackage{amsmath}
\usepackage{natbib}  %
\usepackage{caption} %
\usepackage{subcaption}
\usepackage{algorithm}
\usepackage{algpseudocode}
\usepackage{amsfonts}
\usepackage{multirow}
\usepackage{makecell}
\usepackage[table]{xcolor}

\usepackage[most]{tcolorbox}
\newtcolorbox[list inside=prompt,auto counter]{prompt}[1][]{
    colbacktitle=black!60,
    coltitle=white,
    fontupper=\footnotesize,
    boxsep=5pt,
    left=0pt,
    right=0pt,
    top=0pt,
    bottom=0pt,
    boxrule=1pt,
    #1,
}

\definecolor{DarkGreen}{RGB}{0,140,70}
\definecolor{DarkRed}{RGB}{200,40,40}

\usepackage{newfloat}
\usepackage{listings}
\DeclareCaptionStyle{ruled}{labelfont=normalfont,labelsep=colon,strut=off} %
\floatstyle{ruled}
\newfloat{listing}{tb}{lst}{}
\floatname{listing}{Listing}

\usepackage{booktabs}

\lstdefinestyle{promptstyle}{
    basicstyle=\footnotesize\ttfamily,
    numbers=none,
    frame=single,
    rulecolor=\color{black!35},
    framerule=0.4pt,
    framesep=4pt,
    xleftmargin=0pt,
    framexleftmargin=0pt,
    linewidth=\columnwidth,
    breaklines=true,
    breakatwhitespace=false,
    columns=fullflexible,
    keepspaces=true,
    showstringspaces=false,
    tabsize=2,
    aboveskip=6pt,
    belowskip=6pt,
    captionpos=b
}

\title{QREF: Improving DeepSearch Agents by Aligning Search with Reasoning \\ via Query Refinement and Evidence Filtering}
\author{
    Yujia Chen\textsuperscript{\rm 1,2}\equalcontrib,
    Yuhui Wang\textsuperscript{\rm 1,2}\equalcontrib,
    Junzhe Wang\textsuperscript{\rm 1,2}\equalcontrib,
    Shihan Dou\textsuperscript{\rm 1},
    Di Liang\textsuperscript{\rm 2},
    Yujiong Shen\textsuperscript{\rm 1},\\
    Yuhao Zhou\textsuperscript{\rm 1},
    Zhiheng Xi\textsuperscript{\rm 1},
    Ming Zhang\textsuperscript{\rm 1},
    Senjie Jin\textsuperscript{\rm 1},
    Jiayi Chen\textsuperscript{\rm 1},
    Dingwei Zhu\textsuperscript{\rm 1},\\
    Yajie Yang\textsuperscript{\rm 1},
    Yuan Hua\textsuperscript{\rm 2},
    Chenyao Lu\textsuperscript{\rm 2},
    Tao Gui\textsuperscript{\rm 1}\corresponding,
    Qi Zhang\textsuperscript{\rm 1},
    Xuanjing Huang\textsuperscript{\rm 1}
}
\affiliations{
    \textsuperscript{\rm 1}Fudan University\quad
    \textsuperscript{\rm 2}Tencent\\
    \normalfont\texttt{tgui@fudan.edu.cn}
}

\begin{document}

\maketitle

\begin{abstract}
DeepSearch agents achieve strong performance on complex information-seeking tasks through iterative search and reasoning.
However, they often issue intent-misaligned, under-constrained, or redundant search queries, resulting in poor evidence coverage and accumulating retrieval noise.
To mitigate these limitations, we propose \textbf{QREF}, a plug-and-play \textbf{Q}uery \textbf{R}efiner and \textbf{E}vidence \textbf{F}ilter.
Specifically, the Query Refiner rewrites search queries according to the current reasoning intent while ensuring complementary coverage across each search batch. The Evidence Filter retains evidence relevant to both the current search batch and the reasoning objective, reducing distracting retrieval results.
We initialize the Query Refiner with supervised fine-tuning and optimize it through reinforcement learning based on interactions with a real-world web search environment, reducing ineffective queries.
Comprehensive experiments demonstrate that our approach improves the performance of DeepSearch agents across four challenging benchmarks, achieving average relative Pass@1 improvements of 33.55\% on WebSailor and 11.04\% on TongyiDeepResearch without modifying architectures or retraining their reasoning policies.
Compared with the original DeepSearch backbone without QREF, the RL-trained refiner reduces the fraction of query batches that lack discriminative retrieval anchors by 95\%. Compared with the SFT-trained refiner, it also lowers the filter-based ineffective query ratio.

\end{abstract}

\section{Introduction}
Large language models (LLMs) have demonstrated remarkable capabilities in mathematical reasoning~\cite{guo2025deepseek}, code generation~\cite{li2022competition}, and logical reasoning~\cite{DBLP:conf/nips/Wei0SBIXCLZ22}.
However, hallucinations and unreliable parametric knowledge can still cause factual errors on knowledge-intensive or domain-specific tasks~\cite{DBLP:conf/emnlp/WangWMLGDN24}.
To overcome these limitations, DeepSearch agents interleave reasoning with iterative web search to progressively gather external evidence for complex knowledge-intensive tasks.
Existing methods have improved search performance through enhanced planning, memory mechanisms, and reinforcement learning~\cite{li2025websailor, wu2026webdancer, gao2025beyond, tao2025webshaper}.
However, the quality of intermediate search queries, which strongly influences retrieval effectiveness at each search step, has received much less attention. During search, the agent must translate its evolving reasoning state into one or more executable queries. These queries serve as the interface between reasoning and retrieval, yet their quality is typically optimized only implicitly as part of the overall agent policy.

\begin{figure*}[t]
\centering
\includegraphics[width=\textwidth]{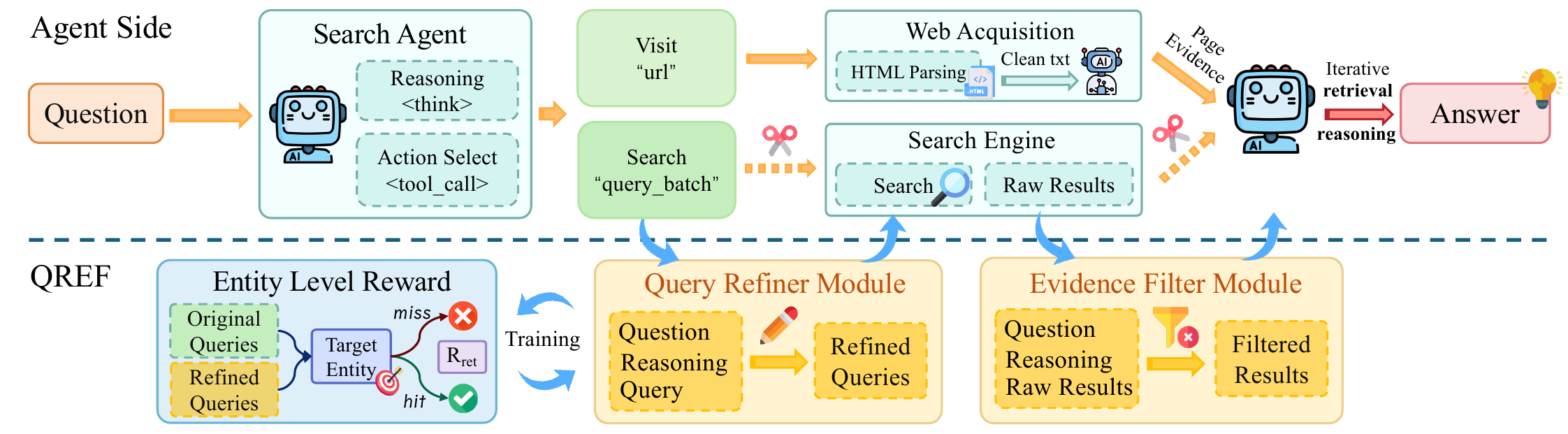} %
\caption{Overview of the QREF integrated into a tool-augmented DeepSearch agent.}
\label{fig:architecture}
\end{figure*}

To characterize query failures in DeepSearch agents, we analyze 300 trajectories generated by WebSailor-32B~\cite{li2025websailor} and identify three recurring failure patterns in intermediate search queries: \textbf{Missing Discriminative Constraints}, \textbf{Low Query Specificity}, and \textbf{Redundant Query Diversification}.
Missing Discriminative Constraints and Low Query Specificity reflect failures of individual queries to adequately capture the retrieval intent implied by the current reasoning state, whereas Redundant Query Diversification reflects insufficient complementary coverage of retrieval intents across queries within the same search batch.
Analysis of 1,992 query batches shows that each of the three failure patterns occurs in more than 20\% of batches, while 47.04\% of batches contain at least one failure.
Moreover, unsuccessful trajectories accumulate nearly three times as many ineffective queries as successful ones on average, with the gap emerging as early as the first search round.
These findings identify intermediate query quality as a critical factor for effective retrieval and multi-hop reasoning.

Recent methods have begun to improve intermediate search behavior through process supervision or retrieval-grounded reinforcement learning~\cite{tao2025webleaper, zheng2025stepsearch, wen2026smartsearch}. Nevertheless, two limitations remain. First, query optimization is generally learned within the search agent's reasoning policy or trajectory-generation process, which limits its reuse with already-trained agents. Second, existing methods typically assess or rewrite queries individually, whereas a DeepSearch step may issue a query batch whose collective coverage matters. Moreover, query reformulation alone cannot prevent real-world search engines from returning irrelevant, redundant, or overly broad evidence. Effective interface optimization therefore requires improving both directions of communication: translating reasoning states into complementary retrieval actions before search, and controlling which retrieved evidence is returned to the reasoning loop afterward.

In this work, we present \textbf{QREF}, a plug-and-play \textbf{Q}uery \textbf{R}efiner and \textbf{E}vidence \textbf{F}ilter for DeepSearch agents. Figure~\ref{fig:architecture} provides an overview of QREF and its integration into the reasoning--search loop of a DeepSearch agent. Specifically, QREF consists of two components. Before retrieval, the \textbf{Query Refiner} rewrites intermediate queries into self-contained, specific, and search-effective formulations while preserving the agent's reasoning intent and ensuring complementary coverage across the search batch. After retrieval, the \textbf{Evidence Filter} retains evidence relevant to both the current query and the reasoning objective, filtering irrelevant, redundant, or overly general search results to reduce retrieval noise. Our module can be integrated into existing DeepSearch agents without modifying their architectures or retraining their reasoning policies.

We first initialize the Query Refiner by supervised fine-tuning (SFT) from Qwen3-8B~\cite{yang2025qwen3technicalreport}. We then further optimize it with Proximal Policy Optimization (PPO)~\cite{DBLP:journals/corr/SchulmanWDRK17} using an entity-level retrieval reward computed by comparing the retrieval outcomes of refined and original queries in real-world web search. Compared with the SFT model, reinforcement learning further improves Pass@1 across seven of eight backbone--benchmark settings and reduces the ineffective-query ratio across all backbone--benchmark settings.

In summary, our main contributions of this work are as follows:
\begin{itemize}
\item We analyze 300 WebSailor-32B trajectories, identifying three recurring failure categories in intermediate queries that each occur in over 20\% of query batches and are strongly associated with lower final task accuracy.

\item We propose QREF, a plug-and-play Query Refiner and Evidence Filter, enabling constraint-preserving, retrieval-oriented query reformulation and evidence filtering without modifying or retraining the underlying agent.

\item We develop a two-stage training strategy for the Query Refiner, combining supervised fine-tuning with PPO under a fine-grained entity-level retrieval reward that directly optimizes evidence recovery in real-world web search.

\item Experiments on four benchmarks and two backbone agents demonstrate consistent improvements in retrieval quality and downstream reasoning, while reducing ineffective and redundant queries.

\end{itemize}

\section{Related Works}

\subsubsection{DeepSearch Agents}
The paradigm of retrieval-augmented language models has evolved from static, single-turn RAG~\cite{lewis2020retrieval} toward autonomous DeepSearch agents that interleave long-horizon planning with iterative interaction with external search engines~\cite{zheng2025deepresearcher,gao2025beyond}. To strengthen the reasoning--retrieval loop, one line of work leverages reinforcement learning to learn search policies directly from real-world retrieval environments~\cite{zheng2025deepresearcher} and to jointly optimize reasoning and search behaviors~\cite{jin2025search,chen2026learning}. A second line scales training to support longer and more effective search trajectories through improved data curation and rollout stability~\cite{team2025tongyi,gao2025beyond,li2025websailor,wu2026webdancer}, while another line augments agents with structured knowledge~\cite{lu2025deepdive}, planning and memory mechanisms~\cite{tao2025webshaper}, or hierarchical retrieval interfaces~\cite{du2026rag}.
Despite these advances, existing methods primarily pursue \emph{agent-level} optimization of reasoning, planning, or search policies, leaving intermediate query generation tightly coupled with the backbone agent and difficult to reuse across already-trained systems.
Our work instead decouples query optimization from the backbone by introducing a backbone-agnostic, plug-and-play refinement module that can be seamlessly integrated into existing DeepSearch agents.

\begin{figure*}[t]
\centering
\begin{subfigure}[t]{0.45\textwidth}
    \centering
    \includegraphics[width=\linewidth]{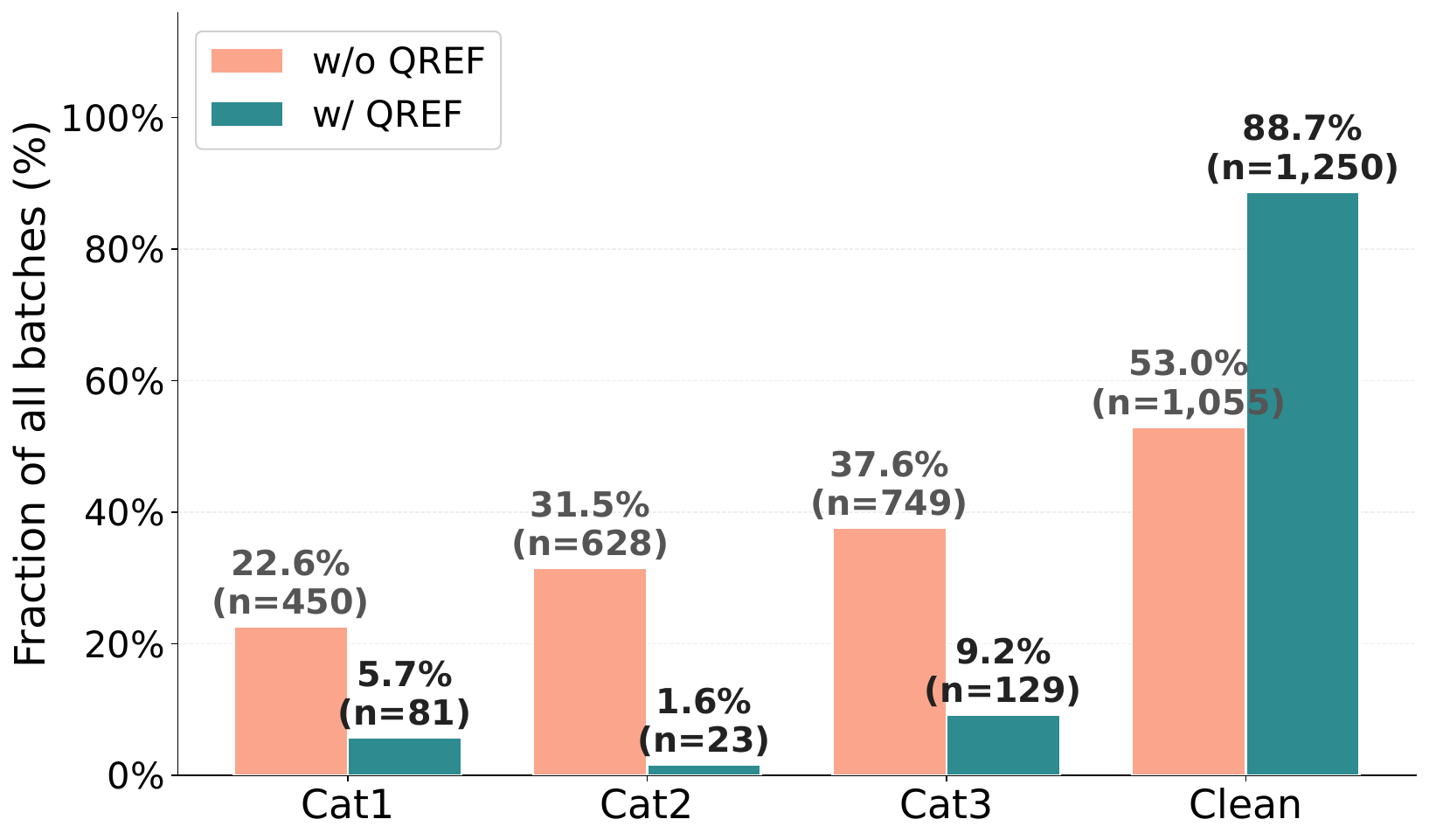}
    \caption{Failure category distribution.}
    \label{fig:failure_bar_orig}
\end{subfigure}
\hfill
\begin{subfigure}[t]{0.45\textwidth}
    \centering
    \includegraphics[width=\linewidth]{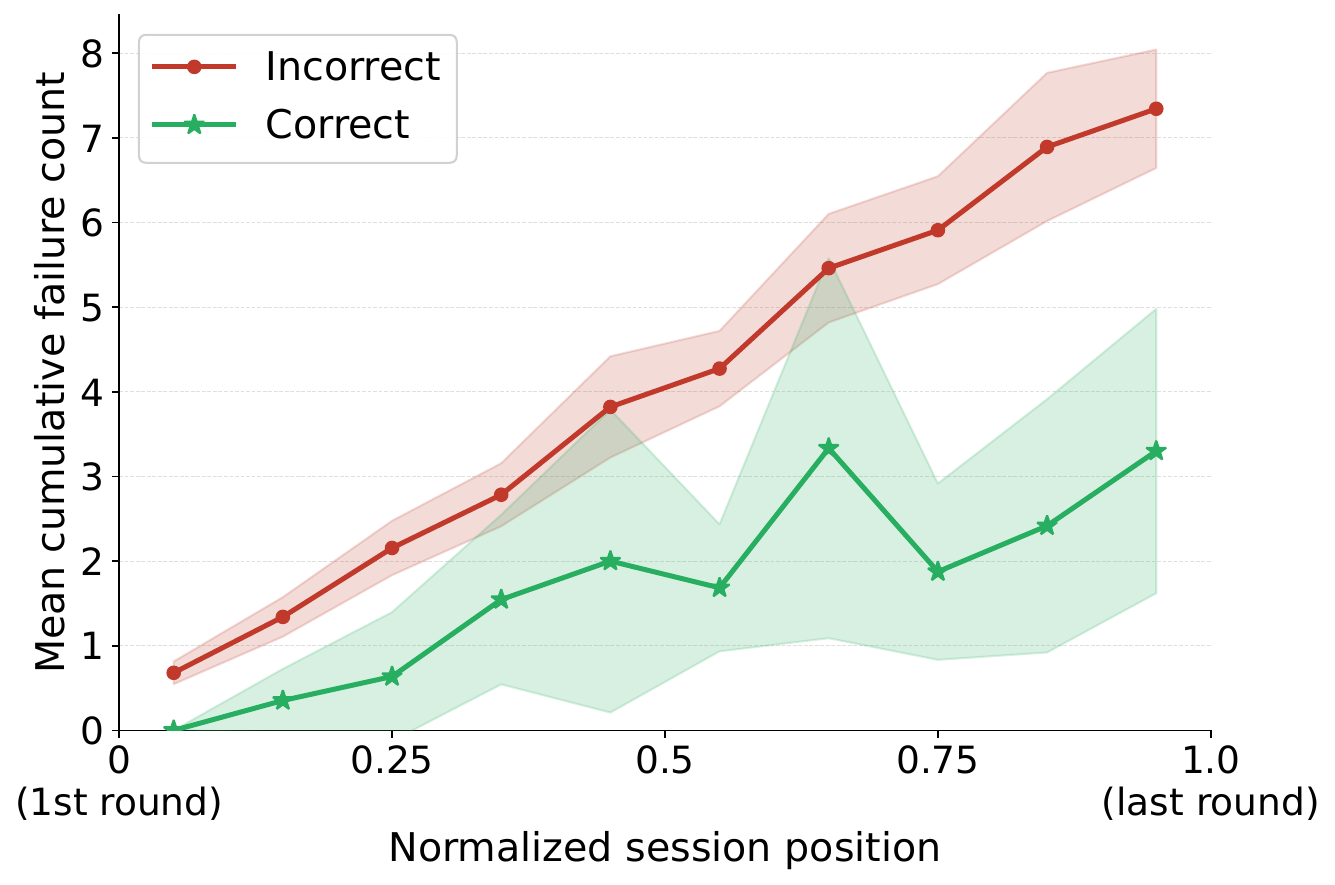}
    \caption{Failure accumulation.}
    \label{fig:accumulation}
\end{subfigure}

\caption{Analysis of ineffective queries on WebSailor. (a) Occurrence rates of the three ineffective-query categories before/after query refinement. (b) Cumulative ineffective-query counts for successful and unsuccessful trajectories.}
\label{fig:failure_analysis}
\end{figure*}

\subsubsection{Query Refinement}
Early query refinement addressed vocabulary mismatch through linguistically motivated expansion~\cite{selvaretnam2016linguistically} and conversational ambiguity through semantic de-contextualization~\cite{jiang2025qagent}. With the advent of LLMs, research shifted toward semantic-level rewriting, where the Rewrite-Retrieve-Read framework established explicit rewriting criteria and trained rewriting policies with reinforcement learning~\cite{ye2023enhancing}, followed by preference alignment and intent-aware optimization~\cite{zhao2025beyond,cao2026multi}. These methods, however, target single-turn retrieval and do not account for queries whose intent evolves with the agent's reasoning state. More recent work extends query optimization to search agents through planning-aware query decomposition~\cite{zhang2025levelrag}, query productivity estimation~\cite{li2026encompass}, process-level supervision of intermediate query generation~\cite{wen2026smartsearch}, and reinforcement learning that directly optimizes retrieval effectiveness~\cite{jiang2025deepretrieval}. In parallel, Over-searching~\cite{xie2026over} identifies redundant low-value retrieval as a common failure mode, and DeepDive~\cite{lu2025deepdive} attributes such failures to the semantic gap between reasoning and executable queries.
However, these methods generally do not explicitly optimize the complementary coverage of query batches or the alignment of retrieved evidence with the current reasoning objective.
In contrast, we cast query optimization as an independent bidirectional reasoning--retrieval interface that optimizes query generation before retrieval and evidence selection afterward, both plug-and-play with respect to the backbone agent.

\section{Diagnosing Ineffective Queries}
\label{sec:query_analysis}
\subsection{Analysis Setup}

To better understand the prevalence, structure, and downstream impact of ineffective intermediate queries, we conduct a focused analysis on 300 randomly sampled WebSailor-32B trajectories from BrowseComp~\cite{DBLP:journals/corr/abs-2504-12516}. These trajectories contain 1,992 search batches and 7,545 individual queries. The category criteria are as follows:
\begin{description}

\item[\textbf{Cat1.}] \textbf{Missing Discriminative Constraints}
The query batch fails to preserve one or more discriminative constraints already identified in the agent's current reasoning.

\item[\textbf{Cat2.}] \textbf{Low Query Specificity}
The query batch lacks sufficiently specific retrieval anchors, causing the search to remain overly broad even if it is semantically relevant.

\item[\textbf{Cat3.}] \textbf{Redundant Query Diversification}
Multiple queries in the batch repeatedly explore the same information need with only superficial wording differences, rather than covering complementary search directions.

\end{description}

Each search batch is labeled with three non-exclusive failure categories, so multiple failure types may co-occur in the same batch.
GPT-4.1 serves as an LLM judge and evaluates the formulation quality of each batch conditioned on the original question and the agent's current reasoning trace. Importantly, the judge assesses the search strategy itself rather than whether the returned results already contain the correct answer. On a 100-batch subset annotated by three human raters, inter-annotator agreement is substantial for all three categories, with Fleiss' $\kappa$ of 0.786, 0.667, and 0.713 for Cat1--Cat3, respectively. GPT-4.1 also agrees well with the human majority vote, achieving $\kappa$/F1 of 0.705/0.839 on Cat1, 0.900/0.950 on Cat2, and 0.770/0.909 on Cat3.

\subsection{Failures Are Prevalent}

We first examine how frequently ineffective query batches occur in WebSailor trajectories. As shown in Figure \ref{fig:failure_bar_orig}, all three failure categories occur in more than 20\% of query batches. Overall, 47.04\% of query batches contain at least one failure, and 86.3\% of complete trajectories contain at least one ineffective batch.

These results show that ineffective intermediate queries are not isolated errors, but a recurring failure pattern in WebSailor. Even a strong search agent frequently fails to convert intermediate reasoning into effective retrieval actions.

\subsection{Failure Categories Co-occur}

We then investigate whether different failure modes occur independently. Based on the binary annotations of the 1,992 query batches, we compute pairwise Pearson correlation coefficients among the three categories.

As illustrated in the supplementary material, all pairwise correlations are positive and statistically significant (all $p < 0.001$), suggesting that these failure behaviors often co-occur rather than arise independently. In particular, \emph{Low Query Specificity} and \emph{Redundant Query Diversification} exhibit the strongest correlation ($r = 0.537$). This pattern suggests that batches with weak retrieval anchors are also more likely to repeat nearly identical search directions instead of covering complementary ones.

\begin{figure*}[th]
\centering
\includegraphics[width=\textwidth]{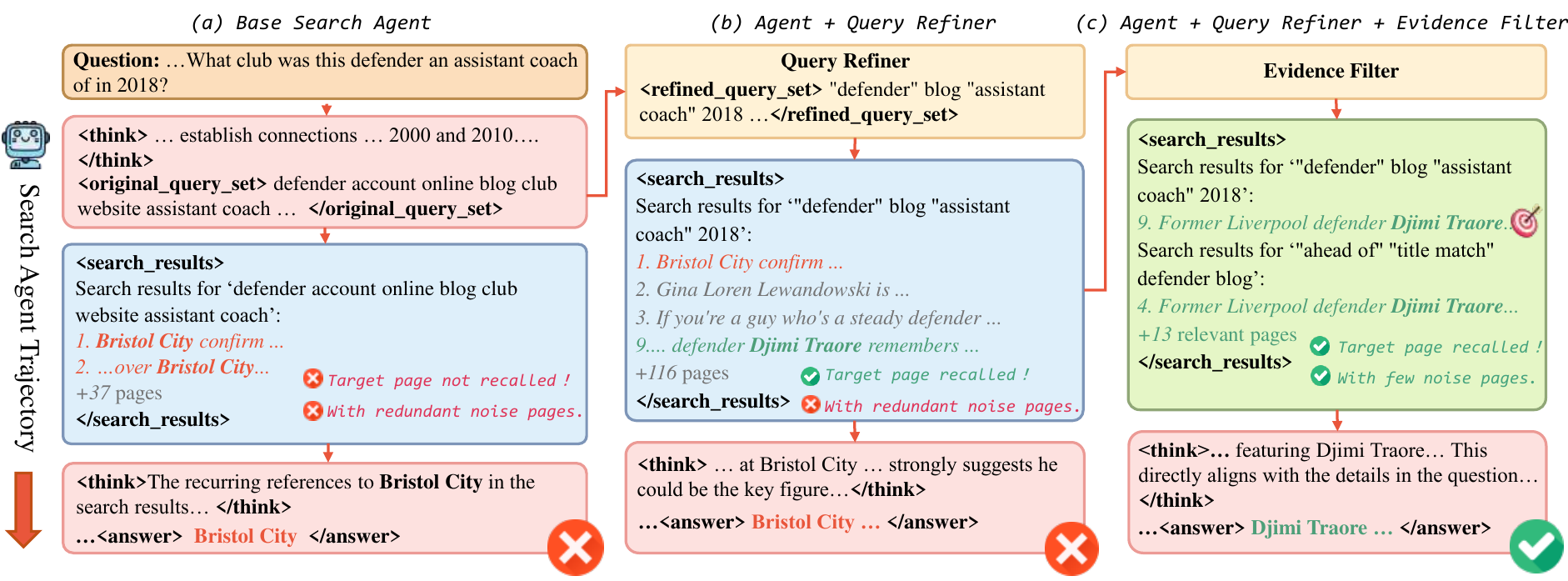} %
\caption{Case study illustrating how the proposed modules refine queries and filter retrieved evidence. (a) Vanilla Search Agent with ineffective intermediate queries. (b) Search Agent with Query Refiner. (c) Search Agent with Query Refiner and Evidence Filter.}
\label{fig:example}
\end{figure*}

\subsection{Failures Are Associated with Lower Final Accuracy}

To quantify the association between ineffective queries and downstream reasoning, we compare the final correctness of trajectories with and without query failures using a $\chi^2$ test. The underlying counts are reported in the supplementary material.

The difference is highly significant ($\chi^2 = 15.997$, $p < 0.001$): trajectories with no ineffective query batches are correct on 26.8\% of examples (11/41), whereas trajectories with at least one ineffective batch are correct on only 6.2\% (16/259). This result indicates that intermediate query quality is strongly associated with end-to-end reasoning performance, rather than merely with retrieval efficiency.

\subsection{Failures Accumulate Over Search Rounds}

We further analyze how ineffective query batches accumulate during multi-step search. Figure~\ref{fig:accumulation} compares cumulative query failures in trajectories that ultimately succeed or fail.

Although both groups exhibit approximately linear growth as search progresses, the two curves diverge from the first search round and continue separating throughout the trajectory. By the end of the search process, unsuccessful trajectories accumulate roughly three times more ineffective batches per trajectory than successful ones.

This observation suggests that ineffective queries rarely occur in isolation. Instead, they accumulate over successive search iterations, progressively increasing retrieval noise, reducing evidence quality, and compounding downstream reasoning errors.

\section{Methodology}

\subsection{QREF}

QREF consists of two components: the Query Refiner and the Evidence Filter. The overall framework is illustrated in Figure~\ref{fig:architecture}.

\paragraph{Query Refiner}
The Query Refiner is an LLM-based rewriting module that takes the original question, the current reasoning trace, and the candidate query batch parsed from the agent's \texttt{<search>} action. It follows four principles. First, \emph{self-containment}: each rewritten query should be interpretable without omitted context or pronouns. Second, \emph{atomic decomposition}: multi-faceted information needs are split into smaller, retrieval-friendly sub-queries. Third, \emph{no answer leakage}: the refiner must not inject external knowledge or hidden answers beyond the current reasoning state. Fourth, \emph{search-engine alignment}: operators such as exact-match quotes, wildcards, negation, and source restrictions (e.g., \texttt{site:}, \texttt{filetype:}) are used only when they improve retrieval precision. Beyond rewriting vague queries, the refiner removes redundant candidates and preserves already effective ones, yielding a query batch that is cleaner and more diverse than the original candidate set.

\paragraph{Evidence Filter}
The Evidence Filter is applied after the search engine returns snippets for each refined query. For every result list \(D_{t,j}^{\mathrm{raw}}\), it jointly considers the original question, the current reasoning trace, and the active refined query \(\tilde{q}_{t,j}\), and predicts which result IDs should be kept. The filter retains evidence that helps identify entities, verify facts, narrow candidates, or advance the current reasoning step, while discarding results that are off-topic, overly general, redundant, or misaligned with the current query intent.

Figure~\ref{fig:example} illustrates this pipeline on a representative BrowseComp question: the Query Refiner introduces the discriminative constraints missing from the agent's original queries, enabling the retrieval of the target evidence, while the Evidence Filter subsequently removes distracting off-topic pages so that the correct entity can be identified.

\subsection{QREF Formulation}
Given a question \(x\), the goal is to produce the final answer \(\hat{y}\) with a tool-augmented web agent. The backbone agent \(\pi_\theta\) maintains an interaction history \(H_t\) and, at each step \(t\), outputs a reasoning trace \(\tau_t\), an action \(a_t \in \{\texttt{search}, \texttt{visit}, \texttt{answer}\}\), and the corresponding action content \(c_t\).

If \(a_t = \texttt{search}\), we parse from \(c_t\) a candidate query batch \(Q_t = \{q_{t,1}, \ldots, q_{t,m_t}\}\). The Query Refiner \(R_\phi\) takes the original question, the current reasoning trace, and \(Q_t\) as input, and outputs
\begin{equation}
o_t^{\mathrm{ref}} = (\rho_t, \tilde{Q}_t) = R_\phi(x, \tau_t, Q_t),
\end{equation}
where \(\rho_t\) is a short rationale and \(\tilde{Q}_t = \{\tilde{q}_{t,1}, \ldots, \tilde{q}_{t,\tilde{m}_t}\}\) is the refined query batch.

The search engine executes each refined query and returns raw results
\begin{equation}
Z_t^{\mathrm{raw}} = \{(\tilde{q}_{t,j}, D_{t,j}^{\mathrm{raw}})\}_{j=1}^{\tilde{m}_t},
\end{equation}
where \(D_{t,j}^{\mathrm{raw}} = \{d_{t,j,k}^{\mathrm{raw}}\}_{k=1}^{n_{t,j}}\) denotes the returned snippets for \(\tilde{q}_{t,j}\), and each snippet contains a title, URL, and short text span. The Evidence Filter \(F_\psi\) evaluates each result list conditioned on \(x\), \(\tau_t\), and the active refined query \(\tilde{q}_{t,j}\), and predicts a kept-index set \(I_{t,j}\). The filtered evidence is
\begin{equation}
Z_t = F_\psi(x, \tau_t, \tilde{Q}_t, Z_t^{\mathrm{raw}})
= \{(\tilde{q}_{t,j}, D_{t,j})\}_{j=1}^{\tilde{m}_t},
\end{equation}
where \(D_{t,j} = \{d_{t,j,k}^{\mathrm{raw}} \mid k \in I_{t,j}\}\).

If \(a_t = \texttt{visit}\), the agent extracts a target URL \(u_t\) from \(c_t\) and obtains page evidence \(v_t = \mathcal{V}(u_t)\). The resulting observation is appended to the interaction history and used in the next step. The process continues until the agent outputs \(a_t = \texttt{answer}\), from which \(\hat{y}\) is extracted.

\begin{algorithm}[t]
\caption{Tool-Augmented Reasoning with the Query Refinement Module}
\label{alg:refine}
\begin{algorithmic}[1]
\Require Question $x$, agent $\pi_\theta$, refiner $R_\phi$, filter $F_\psi$
\Ensure Answer $\hat{y}$

\State $H \gets [m_{\mathrm{sys}}, m_{\mathrm{user}}(x)]$

\While{agent has not produced an answer}
    \State $(\tau_t, a_t, c_t) \gets \mathrm{Parse}(\pi_\theta(H))$

    \If{$a_t = \texttt{search}$}
        \State $Q_t \gets \mathrm{ExtractQueries}(c_t)$
        \State $(\rho_t, \tilde{Q}_t) \gets R_\phi(x, \tau_t, Q_t)$
        \State $Z_t^{\mathrm{raw}} \gets \mathcal{S}(\tilde{Q}_t)$
        \State $Z_t \gets F_\psi(x, \tau_t, \tilde{Q}_t, Z_t^{\mathrm{raw}})$
    \ElsIf{$a_t = \texttt{visit}$}
        \State $u_t \gets \mathrm{ExtractURL}(c_t)$
        \State $Z_t \gets \mathcal{V}(u_t)$
    \ElsIf{$a_t = \texttt{answer}$}
        \State $\hat{y} \gets \mathrm{ExtractAnswer}(c_t)$
        \State \Return $\hat{y}$
    \EndIf

    \State $H \gets H \cup \{(\tau_t, c_t), Z_t\}$
\EndWhile
\end{algorithmic}
\end{algorithm}

\subsection{Cold-Start}
Although LLMs have strong general-purpose query generation ability, direct prompting often yields unstable refinements across reasoning contexts. To obtain a reliable initialization for subsequent reinforcement learning, we first perform SFT on high-quality query refinement demonstrations distilled from GPT-5.2~\cite{openai2025gpt52systemcard}.

Specifically, we construct a query refinement dataset based on DeepResearch-9K~\cite{DBLP:conf/sigir/WuWMHWYZ26}. For each search step, GPT-5.2 rewrites the candidate query batch given the original question, the current reasoning context, and the parsed queries from the agent tool call. Each training instance corresponds to one search turn in an agent trajectory and is represented as a tuple \((x_i, \tau_i, Q_i, o_i^{\mathrm{ref}})\), where \(x_i\) is the original question, \(\tau_i\) is the reasoning context extracted from the immediately preceding \texttt{<think>} block, \(Q_i\) is the candidate query batch parsed from the corresponding \texttt{<tool\_call>} search action, and \(o_i^{\mathrm{ref}} = (\rho_i, \tilde{Q}_i)\) is the GPT-5.2 supervision target containing a concise rationale and the refined query batch.

The query refinement model is optimized using the standard autoregressive language modeling objective over the serialized target output.
This supervised initialization teaches the model effective query refinement behavior and provides a strong starting point for subsequent reinforcement learning. Training on eight NVIDIA H20 GPUs takes approximately two hours.

\subsection{Reinforcement Learning}

Although supervised fine-tuning enables the model to imitate high-quality query refinement behaviors, it primarily learns the surface patterns of rewriting. It does not receive fine-grained rewards from complex and noisy web-search interactions, nor does it explicitly optimize refinement strategies for real-world retrieval or downstream retrieval effectiveness.
Therefore, we further optimize the query refinement policy through reinforcement learning on the InfoSeek dataset~\cite{xia2025opendatasynthesisdeep} using PPO. Training on eight NVIDIA H20 GPUs takes approximately five hours and uses a total of about 100,000 queries.

At each search step, the query refinement model receives the original question, the current reasoning context, and the candidate query batch as input, and generates a refined query batch $\tilde{Q}$ that is subsequently issued to the search engine. The retrieval outcomes of $\tilde{Q}$ are compared against those of the original queries to compute a retrieval-grounded reward, encouraging the model to produce refined queries that recover more of the required evidence than the agent's own queries.

\paragraph{Entity-Level Retrieval Reward}
Instead of relying on a coarse task-success signal, we design a fine-grained reward that directly reflects how well the refined queries retrieve the evidence needed by the current reasoning step. For each training instance, we annotate a set of expected entities $\mathcal{E}$ that the agent must recover from the web at this step. Both the original query batch $Q$ and the refined query batch $\tilde{Q}$ are executed against the search engine, and for each entity $e\in\mathcal{E}$ we determine whether it is \emph{hit} by a query batch.

Concretely, let $\mathcal{R}(q)$ denote the top-$K$ retrieved snippets of query $q$. A single query $q$ is said to hit entity $e$ if a sufficient fraction of its results mention $e$:
\begin{equation}
\mathrm{hit}(q, e)=\mathbb{I}\!\left[
\sum_{r\in\mathcal{R}(q)}\mathbb{I}[e\preceq r]
\;\geq\;
\max\!\left(1,\lfloor \gamma\,|\mathcal{R}(q)|\rfloor\right)
\right],
\end{equation}
where $e\preceq r$ denotes a normalized match of $e$ in snippet $r$, and $\gamma$ is a hit-ratio threshold. A query batch \(Q_{\mathrm{batch}}\) is said to hit \(e\) if at least one
\(q\in Q_{\mathrm{batch}}\) hits \(e\). We denote by $h_Q(e)$ and $h_{\tilde{Q}}(e)$ the hit indicators of the original and refined query batches, respectively.

\paragraph{Reward Formulation}
Based on $(h_Q(e), h_{\tilde{Q}}(e))$, each expected entity falls into one of four disjoint categories, capturing how query refinement changes retrieval coverage:
\begin{itemize}
\item \textbf{Recovered} ($h_Q(e){=}0, h_{\tilde{Q}}(e){=}1$, count $N_{nh}$): the refined queries retrieve an entity that the original queries miss.
\item \textbf{Preserved} ($h_Q(e){=}1, h_{\tilde{Q}}(e){=}1$, count $N_{hh}$): both retrieve the entity.
\item \textbf{Lost} ($h_Q(e){=}1, h_{\tilde{Q}}(e){=}0$, count $N_{hn}$): the refined queries lose an entity originally retrieved.
\item \textbf{Missed} ($h_Q(e){=}0, h_{\tilde{Q}}(e){=}0$, count $N_{nn}$): neither retrieves the entity.
\end{itemize}
The retrieval reward is defined as a weighted per-entity average:
\begin{equation}
R_{\mathrm{ret}}=
\frac{
\alpha_{nh}\,N_{nh}
+\alpha_{hh}\,N_{hh}
-\alpha_{hn}\,N_{hn}
-\alpha_{nn}\,N_{nn}
}{
N_{nh}+N_{hh}+N_{hn}+N_{nn}
},
\end{equation}
where the weights $(\alpha_{nh},\alpha_{hh},\alpha_{hn},\alpha_{nn})$ are set to $(1.0,\,0.1,\,0.8,\,0.2)$.

\begin{table*}[ht]
\centering
\small
\setlength{\tabcolsep}{4pt}
\begin{tabular}{llcccccccccccc}
\toprule
\multirow{2}{*}{Backbone} &
\multirow{2}{*}{Method}
& \multicolumn{3}{c}{BrowseComp}
& \multicolumn{3}{c}{BrowseComp-ZH}
& \multicolumn{3}{c}{GAIA}
& \multicolumn{3}{c}{HLE} \\
\cmidrule(lr){3-5}
\cmidrule(lr){6-8}
\cmidrule(lr){9-11}
\cmidrule(lr){12-14}

&
& P@1 $\uparrow$ & Turns $\downarrow$ & IQR $\downarrow$
& P@1 $\uparrow$ & Turns $\downarrow$ & IQR $\downarrow$
& P@1 $\uparrow$ & Turns $\downarrow$ & IQR $\downarrow$
& P@1 $\uparrow$ & Turns $\downarrow$ & IQR $\downarrow$ \\
\midrule

\rowcolor{gray!15}
\multirow{5}{*}{\shortstack{Tongyi\\DeepResearch}}

& Baseline
&39.89&46.33&-
&39.66&18.10&-
&74.36&10.60&-
&27.61&11.53&- \\

& +Qwen3-8B
&30.65&58.49&59.28
&38.41&28.18&\underline{34.06}
&70.69&11.36&51.01
&25.53&14.01&55.06 \\

& +SFT
&44.39&29.57&62.22
&48.10&15.59&37.22
&74.36&7.77&57.11
&\textbf{28.08}&8.81&59.10 \\

& +RL(Ours)
&\underline{45.62}&\textbf{20.46}&\textbf{36.41}
&\underline{49.48}&\underline{11.50}&\textbf{29.36}
&\underline{77.78}&\underline{7.51}&\textbf{31.21}
&27.73&\underline{6.07}&\textbf{37.25} \\

& +GPT-5.2
&\textbf{46.29}&\underline{22.23}&\underline{52.76}
&\textbf{51.21}&\textbf{10.15}&35.26
&\textbf{79.49}&\textbf{5.97}&\underline{45.15}
&\underline{27.90}&\textbf{5.67}&\underline{45.57} \\

\midrule

\rowcolor{gray!15}
\multirow{5}{*}{WebSailor}

& Baseline
&10.82&\textbf{6.75}&-
&19.03&\underline{4.42}&-
&46.15&\textbf{2.44}&-
&12.09&\textbf{3.63}&- \\

& +Qwen3-8B
&10.43&13.21&\underline{40.95}
&19.03&8.71&\underline{30.22}
&42.74&4.09&\underline{44.80}
&12.33&4.86&\textbf{35.61} \\

& +SFT
&15.56&9.34&71.46
&22.49&4.51&33.35
&\underline{49.57}&3.22&63.74
&12.97&4.71&71.27 \\

& +RL(Ours)
&\underline{17.61}&\underline{7.77}&\textbf{39.33}
&\underline{25.26}&5.02&\textbf{24.25}
&\textbf{52.14}&\underline{3.07}&\textbf{44.16}
&\underline{15.20}&4.44&\underline{50.81} \\

& +GPT-5.2
&\textbf{20.14}&8.03&60.62
&\textbf{25.95}&\textbf{4.39}&31.16
&44.44&3.39&61.64
&\textbf{15.25}&\underline{4.03}&58.71 \\

\bottomrule
\end{tabular}

\caption{
Main results and search efficiency.
P@1 denotes Pass@1.
Turns indicates the average number of search turns for WebSailor and search actions for TongyiDeepResearch.
IQR denotes the ineffective query ratio (\%). \textbf{Bold} and \underline{underlined} numbers denote the best and second-best results, respectively. $\uparrow$ indicates higher is better, and $\downarrow$ indicates lower is better.}

\label{tab:main_results}
\end{table*}

\paragraph{Invalid-Retrieval Penalty}
To prevent the policy from exploiting degenerate rewrites that produce empty or malformed search results, we further impose a hard penalty. Let $\rho_{\emptyset}$ denote the fraction of refined queries in $\tilde{Q}$ that return no results. We clip the reward as
\begin{equation}
R=
\begin{cases}
-1, & \text{if } \tilde{Q} \text{ is unparseable or } \tilde{Q} = \emptyset, \\
-1, & \text{if } R_{\mathrm{ret}}<0 \text{ and } \rho_{\emptyset}\geq \theta_{\emptyset},\\
R_{\mathrm{ret}}, & \text{otherwise}.
\end{cases}
\end{equation}
Additionally, if the ground-truth annotation contains no expected entities or the denominator in $R_{\mathrm{ret}}$ is zero, the reward is set to $0$ so that such instances contribute no gradient signal.

Through this entity-level, retrieval-grounded reward, the policy is directly optimized to enlarge the effective evidence coverage of each search step, rather than merely imitating teacher rewrites or optimizing a coarse end-task signal. In our implementation we use $\gamma{=}0.3$, $K{=}10$, and $\theta_{\emptyset}{=}0.5$.

\section{Experiments}

\subsection{Evaluation Benchmarks}
We evaluate our method on four challenging information-seeking benchmarks, including: (1) \textbf{BrowseComp} (1,266 questions)~\cite{DBLP:journals/corr/abs-2504-12516}, an English web browsing benchmark requiring deep multi-hop web exploration; (2) \textbf{BrowseComp-ZH} ~\cite{DBLP:journals/corr/abs-2504-19314}(289 questions), the Chinese counterpart of BrowseComp requiring multi-hop exploration over Chinese web resources; (3) \textbf{GAIA} (117 questions, text-only subset)~\cite{DBLP:conf/iclr/MialonF0LS24}, a benchmark evaluating multi-step reasoning and tool-assisted problem solving; and (4) \textbf{HLE} (2,158 questions, text-only subset)~\cite{DBLP:journals/corr/abs-2501-14249} a benchmark covering diverse knowledge-intensive reasoning tasks.

\subsection{Backbone Agents}
We evaluate the proposed module on two DeepSearch backbones: WebSailor-32B and TongyiDeepResearch~\cite{tongyidr}. Both agents are served with SGLang and operate in a multi-turn ReAct-style loop that alternates between reasoning, tool use, and answer generation.

\subsection{Auxiliary Models}
The Query Refinement Module is instantiated with four refiner configurations: (1) \textit{+Qwen3-8B}, a prompt-based refiner using a locally deployed open-source LLM; (2) \textit{+GPT-5.2}, a prompt-based refiner using a proprietary API model; (3) \textit{+SFT}, a refiner initialized through supervised fine-tuning; and (4) \textit{+RL}, a refiner further optimized from \textit{+SFT} via reinforcement learning. We use Qwen3-8B as the Evidence Filter. For final-answer evaluation, we follow the backbone models' evaluation pipeline and use benchmark-specific prompts.

\subsection{Implementation Details}

\textbf{Web Search Tools}
The agent interacts with the web through two tools: a search tool powered by the Serper API\footnote{\url{https://serper.dev/}} for query execution and a visit tool powered by the Jina Reader API\footnote{\url{https://jina.ai/reader/}} for webpage retrieval and parsing.

\textbf{Inference Setup}
The backbone agents use a temperature of 0.6, top-$p$ of 0.95, and a maximum context length of 31K tokens. To reduce contamination risk, we apply a rule-based URL blocklist that excludes known benchmark-hosting websites (e.g., \texttt{huggingface}) during retrieval.

\subsection{Main Results}
Table~\ref{tab:main_results} reports \textit{Pass@1} results. Both the trained Query Refiners (SFT and RL) and GPT-5.2 as the Query Refiner consistently improve performance over the vanilla agents across both backbones. On WebSailor, \textit{+RL} achieves the best performance among trained refiners, improving over the baseline by 6.79, 6.23, 5.99, and 3.11 points on BrowseComp, BrowseComp-ZH, GAIA, and HLE, respectively. Similar gains are observed on TongyiDeepResearch, demonstrating the generality of the proposed module across different DeepSearch agents.

Comparing \textit{+SFT} and \textit{+RL}, reinforcement learning outperforms supervised fine-tuning on both backbones, suggesting that retrieval-based rewards provide a more effective optimization signal for query refinement. In contrast, directly applying an off-the-shelf LLM as the refiner is ineffective: Qwen3-8B consistently underperforms the trained refiners and even the baseline on most benchmarks, indicating that general instruction-following ability alone is insufficient.
Although GPT-5.2 achieves the best overall performance, its gains are largely driven by generating substantially more queries, which improves recall at the cost of precision and introduces irrelevant retrievals that reduce search efficiency.

\subsection{Further Analysis}
\subsubsection{Ablation Study}

We ablate the two inference-time components on WebSailor, as shown in Table~\ref{tab:ablation}. Removing either component degrades performance relative to the full model, indicating that both contribute to the overall effectiveness. Using the \textit{Evidence Filter} alone still improves performance on BrowseComp but falls below the baseline on BrowseComp-ZH, suggesting that evidence filtering cannot consistently compensate for low-quality search queries. In contrast, using the \textit{Query Refiner} alone yields consistent improvements over the baseline on both benchmarks. The full model nevertheless achieves the best performance, demonstrating that the two components provide complementary benefits.

\subsubsection{Search Effectiveness}
Let \(Q_{\mathrm{ref}}\) denote the set of all executed refined queries over the evaluation set. A query \(q \in Q_{\mathrm{ref}}\) is considered \emph{ineffective} if, after retrieval and the Evidence Filter, it yields no retained evidence snippet for the current reasoning step. We define the \textbf{ineffective query ratio} as
\begin{equation}
R_{\mathrm{ineff}}=\frac{N^{\mathrm{ineff}}}{N^{\mathrm{all}}},
\end{equation}
where \(N^{\mathrm{all}}\) is the total number of executed refined queries and \(N^{\mathrm{ineff}}\) is the number of queries that produce no retained evidence after filtering.

Table~\ref{tab:main_results} reports the average number of search turns and the ineffective query ratio. On TongyiDeepResearch, query refinement substantially reduces search turns, indicating improved search efficiency, whereas on WebSailor it slightly increases search turns, encouraging the agent to further verify candidate answers through additional exploration. More importantly, the RL-trained refiner achieves a lower ineffective query ratio than the SFT model across all benchmarks while delivering stronger or comparable accuracy, suggesting that reinforcement learning better aligns query refinement with downstream retrieval utility by reducing low-value searches.

\begin{table}[t]
\centering
\small
\setlength{\tabcolsep}{9pt}

\begin{tabular}{lcc}
\toprule
Method &
BrowseComp-ZH &
BrowseComp \\
\midrule

WebSailor
& 19.03
& 10.82 \\

w/o Query Refiner
& 15.22
& \underline{13.51} \\

w/o Evidence Filter
& \underline{22.84}
& 12.88 \\

Full Method
& \textbf{25.26}
& \textbf{17.61} \\

\bottomrule
\end{tabular}

\caption{Ablation study of the proposed modules on WebSailor. \textit{w/o Query Refiner} retains only the Evidence Filter, whereas \textit{w/o Evidence Filter} retains only the Query Refiner.}
\label{tab:ablation}
\end{table}

\subsubsection{Query Quality Analysis}
To further understand the effect of query refinement, we analyze the same 300 questions mentioned in Section~\ref{sec:query_analysis}.
Figure~\ref{fig:failure_bar_orig} compares the occurrence rates of the three failure categories after applying the Query Refinement Module.

As shown in Figure~\ref{fig:failure_bar_orig}, \textit{Low Query Specificity} is reduced by 95\%, indicating that refined queries introduce more discriminative retrieval anchors. \textit{Missing Discriminative Constraints} and \textit{Redundant Query Diversification} are both reduced by approximately 3$\times$, suggesting better preservation of reasoning constraints and more complementary search directions. We calculate the pairwise Intersection over Union (IoU) within query batches for the w/o QREF and w/ QREF settings.
The average query IoU within batches dropped from 0.41 to 0.29, indicating reduced redundancy and increased diversity.

\section{Conclusion}

In this work, we propose \textbf{QREF}, a plug-and-play Query Refiner and Evidence Filter. The Query Refiner rewrites intermediate search queries into self-contained, constraint-aware, and complementary formulations before retrieval, and the Evidence Filter retains only evidence relevant to both the refined query and the reasoning objective. Furthermore, we devise a two-stage training strategy that couples supervised fine-tuning with reinforcement learning under an entity-level retrieval reward in a real-world web search environment.
Our method consistently improves retrieval and reasoning on four benchmarks with two backbones,  delivering improved retrieval quality and downstream reasoning while substantially reducing the proportion of ineffective queries. This work highlights intermediate query quality as a key interface between reasoning and retrieval in DeepSearch agents.

\bibliography{aaai2027}

\clearpage
\appendix
\twocolumn[{\centering\LARGE\bfseries Appendix\par\vspace{1em}}]

\section{Case Study}

Figure \ref{fig:example} presents a representative example illustrating how our two modules progressively improve retrieval quality and downstream reasoning.

Given the question ``Ahead of a title match, a defender who played for one of the teams... What club was this defender an assistant coach of in 2018'', the vanilla DeepSearch agent first generates several vague intermediate queries such as "assistant coach defender blog club website" and "defender account online blog assistant coach". These queries omit critical constraints (e.g., the title match, the defender's identity, and the associated event), causing the search engine to retrieve numerous irrelevant pages centered around Bristol City. Since the target evidence never appears in the retrieved documents, the agent incorrectly infers that Bristol City is the answer.

After applying the Query Refiner, the intermediate queries become substantially more search-oriented. The refiner decomposes the information need into multiple complementary queries while explicitly preserving important constraints, producing queries such as ``"defender" blog "assistant coach" 2018" and ``"ahead of" "title match" defender blog''. As a result, the target page discussing Djimi Traoré is successfully retrieved. However, the retrieved evidence still contains many irrelevant pages because the search results include documents matching only partial query intents. Consequently, although the correct evidence is present, the reasoning process can still be distracted by redundant information and arrive at the same incorrect conclusion.

Finally, the Evidence Filter removes documents that are inconsistent with the current reasoning objective while preserving evidence directly supporting the intermediate query. The remaining context is dominated by pages describing Djimi Traoré, allowing the reasoning process to correctly establish the connection between the defender and his role as an assistant coach in 2018, ultimately producing the correct answer.

This example demonstrates that the two modules address complementary bottlenecks in DeepSearch agents. Query Refinement primarily improves target evidence recall by generating more discriminative and search-effective queries, whereas Evidence Filtering further improves the signal-to-noise ratio of the retrieved context, enabling the downstream reasoning module to utilize the recalled evidence more reliably. Only when both modules are combined can the agent consistently retrieve the correct evidence and reason over it successfully.

\newpage
\section{Prompt Templates}
\label{app:prompts}

This section reports the prompts used by the Query Refiner and
Evidence Filter. The placeholders enclosed in angle brackets are
replaced with the corresponding trajectory information at inference
time.

\subsection{Query Refiner Prompt}
\label{app:query-refiner-prompt}

\begin{prompt}[title={query-refiner-prompt}]{
    You are a professional assistant skilled at rewriting complex or unclear questions into simpler, precise, and search-engine-friendly sub-questions.

Your task is to generate a set of search-engine-friendly queries based on the provided input context (Question, Current Reasoning, and Candidate Queries).

OPERATIONAL GUIDELINES:\\
1. \textbf{Self-Containment}: \\Each query must be independent and understandable without the original context. Avoid all demonstrative pronouns (e.g., "this", "it").\\
2. \textbf{Atomic Decomposition}: \\Break down multi-faceted information needs into distinct, atomic sub-questions.\\
3. \textbf{No Spoilers /Inference}: \\Strictly do NOT leak potential answers from the reasoning chain into the queries. Do not use internal knowledge to bridge gaps; let the search engine do the work.\\
4. \textbf{Search Syntax Mastery}:\\
   - Use `""' for exact matching to precisely search for specific entities.\\
   - Use the asterisk `*' as a wildcard for unknown terms in a phrase with `""'(e.g., `"best * architecture"').\\
   - Use `-' to disambiguate (e.g., `mercury -planet' if searching for the element).\\
   - Use `site:' or `filetype:' if the reasoning suggests a need for official documentation or specific formats.\\
5. \textbf{Evolutionary Filtering}: \\Analyze the "Current candidate query set". Retain high-quality queries, refine vague ones, and eliminate redundant or overlapping ones.

CONSTRAINTS:\\
- \textbf{Conciseness}: \\No conversational filler ("Search for...", "Find information about..."). Use information-dense keywords.\\
- \textbf{No Hallucination}: \\Do not invent entities or facts not contained in the input.\\
- \textbf{Output Integrity}: \\You must output ONLY a valid JSON object.

INPUT CONTEXT:\\
- Question: \{{QUESTION}\}\\
- Current reasoning: \{{THOUGHT}\}\\
- Current candidate query set: \{{QUERY}\}

OUTPUT FORMAT:\\
\{\\
  "reasoning": "Explain the strategy for the refinement, specifically why certain operators were chosen or why the candidate set was modified.",\\
  "refined\_query\_set": ["sub-query-1", "sub-query-2"]\\
\}
}\end{prompt}

% \begin{lstlisting}[
%     style=promptstyle,
%     caption={Prompt used by the Query Refiner.},
%     label={lst:query-refiner-prompt}
% ]
% \end{lstlisting}

\subsection{Evidence Filter Prompt}
\label{app:evidence-filter-prompt}
\begin{prompt}[title={evidence-filter-prompt}]{
You are a search-result filtering expert for a DeepSearch agent.

Your task is to remove noisy, irrelevant search results and retain only those that genuinely help advance the agent's current information need.

Judge usefulness strictly based on:\\
- The original question\\
- The agent's current reasoning and search intent\\
- The current search query

Keep results that can:\\
- Identify key entities,\\
- Align with search intent,\\
- Verify critical facts,\\
- Narrow down candidates,\\
- Or directly advance reasoning.

Discard results that are:\\
- Off-topic\\
- Overly general\\
- Background-only\\
- Redundant\\
- Not aligned with the current query intent

[Input]

Original Question:
{question}

Current Reasoning (for internal understanding only):
{thought}

Current Search Query:
{query}

Search Results:
{results}

[Output Format]

Output ONLY the following JSON, without any extra text:

\{
  "kept\_ids": [1, 2, ...]
\}
    }
\end{prompt}
% \begin{lstlisting}[
%     style=promptstyle,
%     caption={Prompt used by the Evidence Filter.},
%     label={lst:evidence-filter-prompt}
% ]

% \end{lstlisting}

\section{Training Details}
\label{app:training-details}

We summarize the main training hyperparameters used for supervised fine-tuning
(SFT) and reinforcement learning (RL). Machine-specific settings, such as
internal paths, proxy variables, cluster startup commands, and system memory
configuration, are omitted because they do not affect the training objective.

\subsection{Supervised Fine-Tuning}
\label{app:sft-details}

We first perform full-parameter SFT on the Hunyuan backbone. The model is
trained for one epoch with a maximum sequence length of 20,480 tokens using
bfloat16 precision. We use a learning rate of $1\times10^{-5}$ with a cosine
schedule and a warmup ratio of 0.1. DeepSpeed ZeRO-3 with parameter and optimizer
offloading is used to reduce memory consumption. FlashAttention-2, the Liger
kernel, and gradient checkpointing are enabled for efficient long-context
training. Table~\ref{tab:sft-hyperparameters} summarizes the main settings.

\subsection{Reinforcement Learning}
\label{app:rl-details}

\begin{table}[h]
\centering
\small
\begin{tabular}{@{}p{0.38\columnwidth}p{0.56\columnwidth}@{}}
\toprule
Setting & Value \\
\midrule
Fine-tuning type & Full-parameter SFT \\
Maximum length & 20,480 tokens \\
Batch size & 2 per device \\
Gradient accumulation & 2 steps \\
Learning rate & $1\times10^{-5}$ \\
Training epochs & 1 \\
Scheduler & Cosine, 0.1 warmup ratio \\
Precision & bfloat16 \\
Distributed optimization & DeepSpeed ZeRO-3 offload \\
Efficiency features & FlashAttention-2, Liger kernel, gradient checkpointing \\
\bottomrule
\end{tabular}
\caption{Main hyperparameters used for supervised fine-tuning.}
\label{tab:sft-hyperparameters}
\end{table}

Starting from the SFT checkpoint, we further optimize the model with proximal
policy optimization (PPO) using the VERL training framework. Advantages are
estimated with generalized advantage estimation (GAE), and rewards are computed
by a task-specific reward function. Training uses a batch size of 64, with a
maximum prompt length of 2,048 tokens and a maximum response length of 2,048
tokens. The actor and critic learning rates are $1\times10^{-6}$ and
$1\times10^{-5}$, respectively. We generate one rollout per prompt and train for
one epoch. KL regularization is applied through the reward, while an additional
actor KL loss is disabled. Table~\ref{tab:rl-hyperparameters} lists the main RL
settings.

\begin{table}[h]
\centering
\small
\begin{tabular}{@{}p{0.38\columnwidth}p{0.56\columnwidth}@{}}
\toprule
Setting & Value \\
\midrule
RL algorithm & PPO with GAE \\
Training batch size & 64 \\
Prompt / response length & 2,048 / 2,048 tokens \\
Actor / critic learning rate & $1\times10^{-6}$ / $1\times10^{-5}$ \\
PPO mini-batch size & 64 \\
Actor / critic micro-batch & 1 / 2 per GPU \\
Rollouts per prompt & 1 \\
Training epochs & 1 \\
Entropy coefficient & 0 \\
KL configuration & KL in reward enabled; actor KL loss disabled \\
Parallelism & 1 node, 8 GPUs; actor TP$=2$, rollout TP$=1$ \\
Pipeline / context parallelism & PP$=1$, CP$=1$ \\
Batching and token budget & Dynamic batching; 4,096 tokens per GPU for actor, critic, reference, and rollout log-prob computation \\
Rollout engine & vLLM; memory utilization 0.7; 15,000 maximum batched tokens \\
Memory optimization & Actor, critic, reference, gradient, and optimizer offloading \\
\bottomrule
\end{tabular}
\caption{Main hyperparameters used for PPO training. TP, PP, and CP denote tensor, pipeline, and context parallelism, respectively.}
\label{tab:rl-hyperparameters}
\end{table}

\newpage
\section{Additional Results and Failure Analysis}
\label{app:additional-results}

This section provides complementary results for the WebSailor-based agent and
additional analyses of ineffective query generation. We first report performance
under three-attempt evaluation and then examine how ineffective query batches
relate to trajectory correctness and to one another.

\subsection{WebSailor Results under Three-Attempt Evaluation}
\label{app:websailor-pass3}

Table~\ref{tab:sailor_pass3} reports the performance of WebSailor equipped with
the Query Refinement Module on BrowseComp-English (BC-EN), BrowseComp-Chinese
(BC-ZH), Humanity's Last Exam (HLE), and GAIA. Across all four benchmarks,
Pass@3 is higher than Avg@3, indicating that allowing multiple attempts improves
the probability of obtaining at least one successful trajectory. The largest
Pass@3 values are observed on GAIA, followed by BC-ZH, while HLE remains the
most challenging benchmark under both metrics.

\begin{table}[h]
\centering
\small
\setlength{\tabcolsep}{8pt}
\begin{tabular}{lcccc}
\toprule
Method & BC-EN & BC-ZH & HLE & GAIA \\
\midrule
Avg@3
& 18.07 & 23.99 & 15.06 & 48.43 \\
Pass@3
& 30.15 & 41.18 & 25.81 & 65.51 \\
\bottomrule
\end{tabular}
\caption{Additional WebSailor results with the Query Refinement Module, reported using \textit{Avg@3} and \textit{Pass@3}.}
\label{tab:sailor_pass3}
\end{table}

\subsection{Ineffective Query Batches and Trajectory Correctness}
\label{app:failure-batch-analysis}

To quantify the relationship between ineffective query generation and final
answer correctness, we divide trajectories according to whether they contain at
least one ineffective query batch. Table~\ref{tab:chi_square_failure} gives the
contingency counts used in the $\chi^2$ analysis. Among trajectories without a
failure batch, 11 of 41 are correct, whereas only 16 of 259 trajectories with at
least one failure batch are correct. This difference shows a strong association
between ineffective intermediate search behavior and unsuccessful final
trajectories, while not by itself establishing a causal relationship.

\begin{table}[!htbp]
\centering
\small
\begin{tabular}{lccc}
\toprule
Trajectory Group & Incorrect & Correct & Total \\
\midrule
No failure batches & 30 & 11 & 41 \\
At least one failure batch & 243 & 16 & 259 \\
\midrule
Total & 273 & 27 & 300 \\
\bottomrule
\end{tabular}
\caption{Contingency table for the $\chi^2$ analysis of trajectory correctness versus the presence of ineffective query batches.}
\label{tab:chi_square_failure}
\end{table}

\subsection{Relationships among Ineffective-Query Categories}
\label{app:ineffective-query-correlations}

\begin{figure}[!htbp]
\centering
\includegraphics[width=\columnwidth]{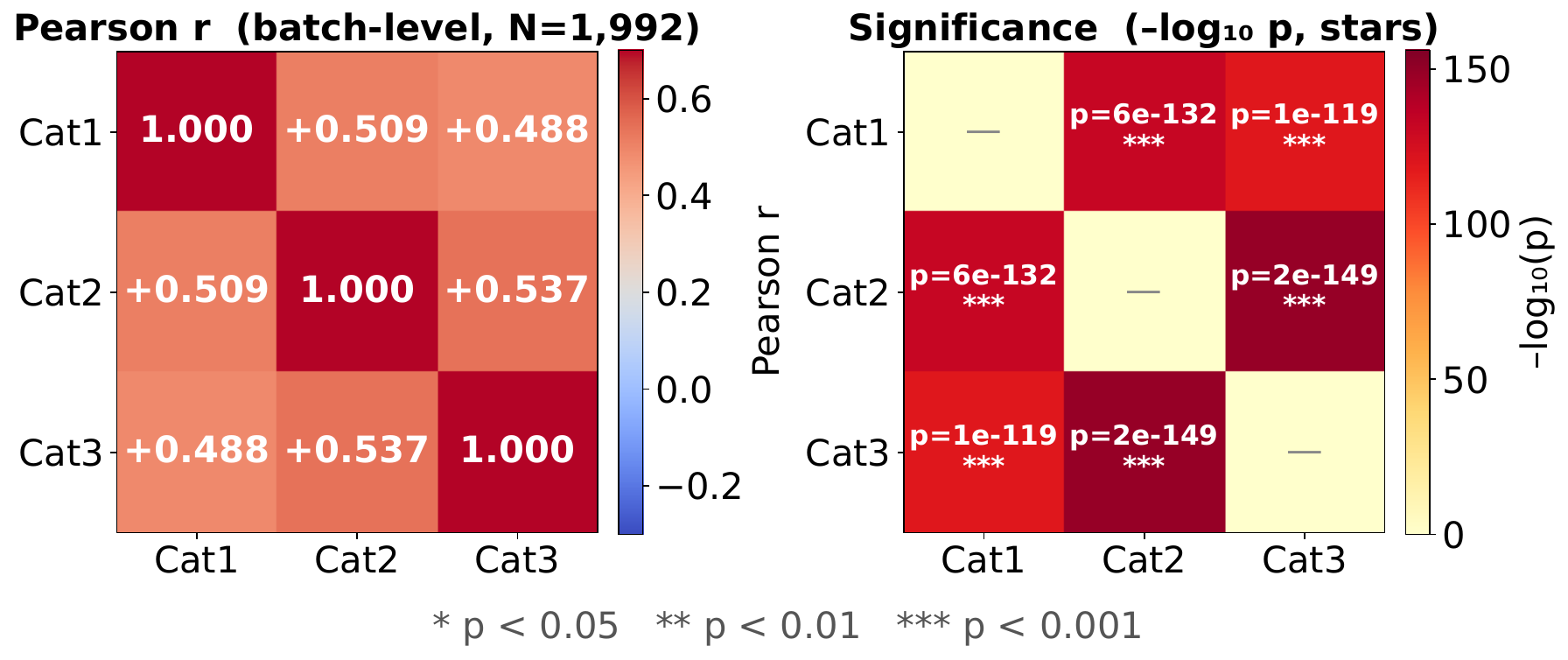} %
\caption{Pairwise Pearson correlation matrix for the three ineffective-query categories. The heatmap summarizes the extent to which different query-failure modes co-occur across trajectories.}
\label{fig:pearson}
\end{figure}

We compute pairwise Pearson correlations among the three failure modes using annotations from 1,992 query batches. As shown in Figure ~\ref{fig:pearson}, all correlations are positive and statistically significant ($p < 0.001$), indicating that the failure modes tend to co-occur. The strongest correlation is observed between \emph{Low Query Specificity} and \emph{Redundant Query Diversification} ($r = 0.537$).

\end{document}